\documentclass{article}
\usepackage{acra}
\usepackage{siunitx}
\usepackage{graphicx}
\usepackage{multirow}
\usepackage{adjustbox}
\usepackage{amsmath}
\usepackage{url}
\usepackage{hyperref}
\usepackage{algorithm}
\usepackage{algpseudocode}
\usepackage[table,xcdraw]{xcolor}

\title{Seeing the Fruit for the Leaves: Towards Automated Apple Fruitlet Thinning}
\author{
  Ans Qureshi$^{1*}$, Neville Loh$^1$, Young Min Kwon$^1$, David Smith$^1$\\ \textbf{Trevor Gee$^1$, Oliver Bachelor$^2$, Josh McCulloch$^2$, Mahla Nejati$^1$, JongYoon Lim$^1$}\\
  \textbf{Richard Green$^2$, Ho Seok Ahn$^1$, Bruce MacDonald$^1$, Henry Williams$^{1**}$}\\
  Centre for Automation and Robotic Engineering Science, The University of Auckland, NZ$^1$\\
  Department of Computer Science and Software Engineering, University of Canterbury, NZ$^2$\\
  \texttt{aqur476@aucklanduni.ac.nz$^*$, henry.williams@auckland.ac.nz$^{**}$}
}

\begin{document}

\maketitle
\begin{abstract}
    Following a global trend, the lack of reliable access to skilled labour is causing critical issues for the effective management of apple orchards.
    One of the primary challenges is maintaining skilled human operators capable of making precise fruitlet thinning decisions.
    Thinning requires accurately measuring the true crop load for individual apple trees to provide optimal thinning decisions on an individual basis. 
    A challenging task due to the dense foliage obscuring the fruitlets within the tree structure. 
    This paper presents the initial design, implementation, and evaluation details of the vision system for an automatic apple fruitlet thinning robot to meet this need.
    The platform consists of a UR5 robotic arm and stereo cameras which enable it to look around the  leaves to map the precise number and size of the fruitlets on the apple branches. 
    We show that this platform can measure the fruitlet load on the apple tree to with 84\% accuracy in a real-world commercial apple orchard while being 87\% precise.

\end{abstract}

\section{Introduction}
    The New Zealand apple and pear industry has a current export value of \$917m and is expected to reach \$1 billion by 2022 \cite{applesNZ2022}. 
    The industry's current trajectory will have it at \$2 billion by 2030.
    Following a similar global trend, the projected industry expansion could require a larger seasonal workforce which will need to be trained in how to complete skilled, seasonal tasks such as apple fruitlet thinning.
    Fruitlet thinning is a critical tool used by apple growers to regulate crop load (fruitlet count) on the trees to maximise the quality of the apples produced.

    Thinning the apple crop is a vital part of the growing cycle to manage the crop load on each tree. 
    Thinning involves removing apple fruitlets by hand from the apple trees. 
    This needs to happen to get the right colour and size for each apple variety. 
    This job requires good attention to detail as the thinner has to assess and balance the size, shape, colour, clustering, and the total number of apples across the tree. 
    Providing optimal spacing and distribution of the fruitlets across the tree will maximise the number of high-quality apples produced.

    Current thinning approaches will estimate the current load on the trees in the orchard by manually counting the number of fruitlets and clusters on a representative number of trees.
    The tree's desired load is calculated for the given orchard based on expected weather conditions, general tree conditions, and yield predictive models.
    Orchard managers will then determine the size and number of clusters each tree can manage to achieve the desired yield.
    Thinners are then instructed to reduce the clusters on the trees to meet these specifications, presuming that this will achieve the desired count for each tree.
    
    This is typically achieved by removing the damaged and undersized fruitlets first. 
    Next, the least mature fruitlets from each cluster will be removed to reduce the size of clusters, ideally leaving only clusters of two or three based on the desired yield. 
    Finally, the thinner will attempt to balance the load across the tree to provide adequate spacing between the fruitlets to grow to the desired size \footnote{Several variations to this method exist between orchard management practices but follow this general concept.}.

    These estimated approaches work on average across the entire orchard, but an individual tree will likely end up under or overloaded. 
    The count and desired load are presumed to represent all the trees in the entire orchard.  
    Ideally, each tree's current load and capacity would accurately be determined individually.
    Specific thinning rules for the given tree can then be determined to maximise its yield and quality. 
    Our Horticultural collaborators have advised that the tree level thinning approach could translate to a 10\%-30\% increase in quality.
    For the New Zealand apple industry, this would translate to an additional \$95million pa (assuming 10\% increased yield produces an additional 48,400Tpa nationally and an average apple price \$1,971/T).

\begin{figure}[htb]
        \centering
        \includegraphics[width=\linewidth]{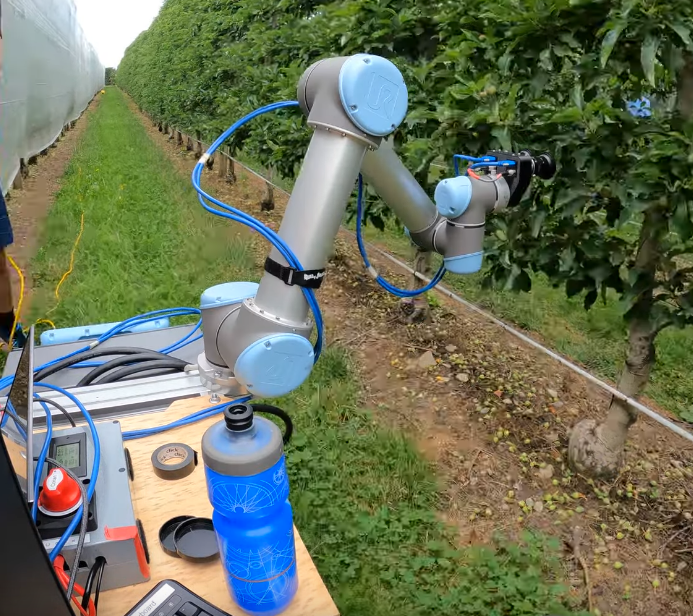}
        \caption{Robotic platform used to scan the apple trees branch by branch in a commercial orchard.}
        \label{fig:scanning_rig}
    \end{figure}

    Our work aims to develop a robotic platform capable of thinning apple trees based on their specific load and yield capacities.
    It is worth noting that this work specifically targets 2D apple tree structures where this approach is significantly more feasible. 
    This paper presents the initial design, implementation, and evaluation details of the vision system of this robotic platform.
    The platform consists of a UR5 robotic arm and stereo cameras which enable it to look around the apple branches to measure the number and size of the fruitlets along the tree branches. 
    Figure \ref{fig:scanning_rig} presents a photo of the scanning platform prototype and Section \ref{sec:videos} provides videos of it operating.

    The performance of the thinning platform's vision system has been measured through a comprehensive and realistic field trial in a commercial environment with quantified performance metrics compared to ground truth measurements.
    Before the field trials, no modifications to or thinning of the orchard were conducted, with the thinner operating as required in commercial practice.
    The results of these field trials are presented, and the implications for further platform development are discussed.
    
    The New Zealand Government and industry investment jointly funded the project to develop this platform.
    It is a collaboration between the Universities of Auckland, Waikato, Canterbury, and Otago, along with Plant and Food Research, and Lincoln Agritech ltd\footnote{MBIE contract UOAX1810 Decision Automation for Orchards and Vineyard “Human Assist”}.
    
\section{Related Work}
    Skill shortages worldwide have led to significant demand for automation of various tasks in orchards and vineyards.
    Work in precision agriculture has been conducted for several decades but has taken off more recently with significant advances in robotics and machine vision.
    Most of the automation work in this area has focused on harvesting \cite{Silwal2017,williams2019robotic,williams2020improvements}, or pollination \cite{amador2017,ohi2018design,williams2020autonomous,williams2021evaluating} tasks to supplement the workforce directly.
    However, yield estimation is a critical area for planning the overall workload throughout the season, and accurate crop yield measurements can provide key information for predicting workload throughout the year.
    
    Over the last two decades, extensive work has been conducted to develop reliable means of predicting or measuring a crop's yield \cite{sagar2018agriculture,koirala2019deep,van2020crop}. 
    This work falls into two main categories for estimating yield, seasonal model-based predictions \cite{basso2019seasonal} and direct counting of fruit on the canopy \cite{koirala2019deep}. 
    
    The most common approach to crop yield estimation attempts to use big data to produce growth models to predict the output of a given crop \cite{sagar2018agriculture}.
    These approaches target large-scale farming with crops such as rice, soybeans, and corn. 
    Measurements such as agriculture soils, climate conditions, weather forecasting, soil quality, and yearly crop yields are gathered to produce models that can predict the yield of a given crop for the new year \cite{basso2019seasonal}.  
    For example, the work in \cite{prasad2006crop} produced a model capable of predicting the soybean yields across Iowa in the United States to an $R^2$ value of 0.86.
    Recent years have seen the extensive use of machine learning to produce these models, with the advent of remote and satellite sensing providing abundant information \cite{ferencz2004crop,awad2019toward}.
    These approaches are appropriate for large-scale farming operations but lack the fidelity to predict individual vineyards or farms accurately. 
    
    Recent advances in machine learning for computer vision have led to extensive work aiming to predict crop yield by counting the individual fruit in an orchard or vineyard \cite{koirala2019deep}. 
    Uses for machine vision detection of fruit in images of tree canopies include estimation of fruit number per tree (‘load’), in-field fruit sizing, and automated harvest. 
    Estimation of fruit size together with fruit number allows estimation of fruit weight (‘yield’) per orchard.
    This work has been applied to a wide range of crops including apples \cite{wang2013automated,wang2017tree}, oranges \cite{chen2017counting}, mangoes \cite{payne2013estimation}, and kiwifruit \cite{massah2021design,mekhalfi2020vision}. 

    Work by \cite{wang2017tree} measured the average size of mangoes using a single snapshot of a mango tree with an RGBD camera.  
    Their machine vision system utilising machine learning was capable of measuring the size and shape of the mangoes too, within an RMSE of \SI{4.9}{\mm} and \SI{4.3}{\mm} for fruit length and width.
    In the case of mangoes, the fruit visible on the tree's outer edge is representative of the fruit obscured within the tree as they grow relatively uniformly.  
    As such, the method does not require the detection of every fruit on the tree to estimate the fruit's size overall.
    A tree fruit load estimate relies on assessing the total number of fruit per tree, not the number of fruit visible in an image.

    One approach to estimating fruit number per tree involves increasing the number of viewpoints around the tree. 
    The idea is to visualise all fruit on the tree by looking around obstructions such as branches or leaves. 
    These multiple view approaches are computationally complex and require precise object-tracking algorithms. 
    Tracking can be challenging as fruit are visually similar, making determining which fruit is which between views complex. 
    
    \cite{payne2013estimation} reported a higher correlation between image counts and human counts of total fruit on the tree based on a summation of counts from images taken on four sides of each canopy compared to single images. 
    This form of multiple imaging of one canopy can result in over-estimation of fruit load by multiple counting of the same fruit at different angles. 
    Further work sought to resolve this issue by adding 3D spatial information to register fruit between frames \cite{wang2013automated,song2014automatic,stein2016image}. 
    
    The work by \cite{stein2016image} was the most recent and complete of these approaches.
    They used a multiple view approach of 37 images around a mango tree from a moving platform.
    Fruit number per tree was estimated based on an epipolar projection approach with fruit tracking using trajectory data (camera pose) provided by the navigation system and LIDAR tracking the trees. 
    They demonstrated an error rate of 1.36\% over 16 individual trees when comparing their count against the harvested count.
    However, the evaluation did presume that any double counting of the fruit was balanced by the non-detection of any hidden fruit (obstructed) fruit. 
    The validation that the mangoes counted by the system were true to the tree was not measured. 

    Overall, prior approaches still cannot see all the fruit for the leaves; they rely on an overall estimation of the number of fruits on the given canopy or tree. 
    In some cases, statistical adjustments are made to approximate the true count from the presumed partial count. 
    Our work towards automated thinning requires counting the precise number of fruitlets and finding their locations along each branch of the tree structure. 
    
\section{Robotic Platform}
\begin{figure}[htb]
        \centering
        \includegraphics[width=\linewidth]{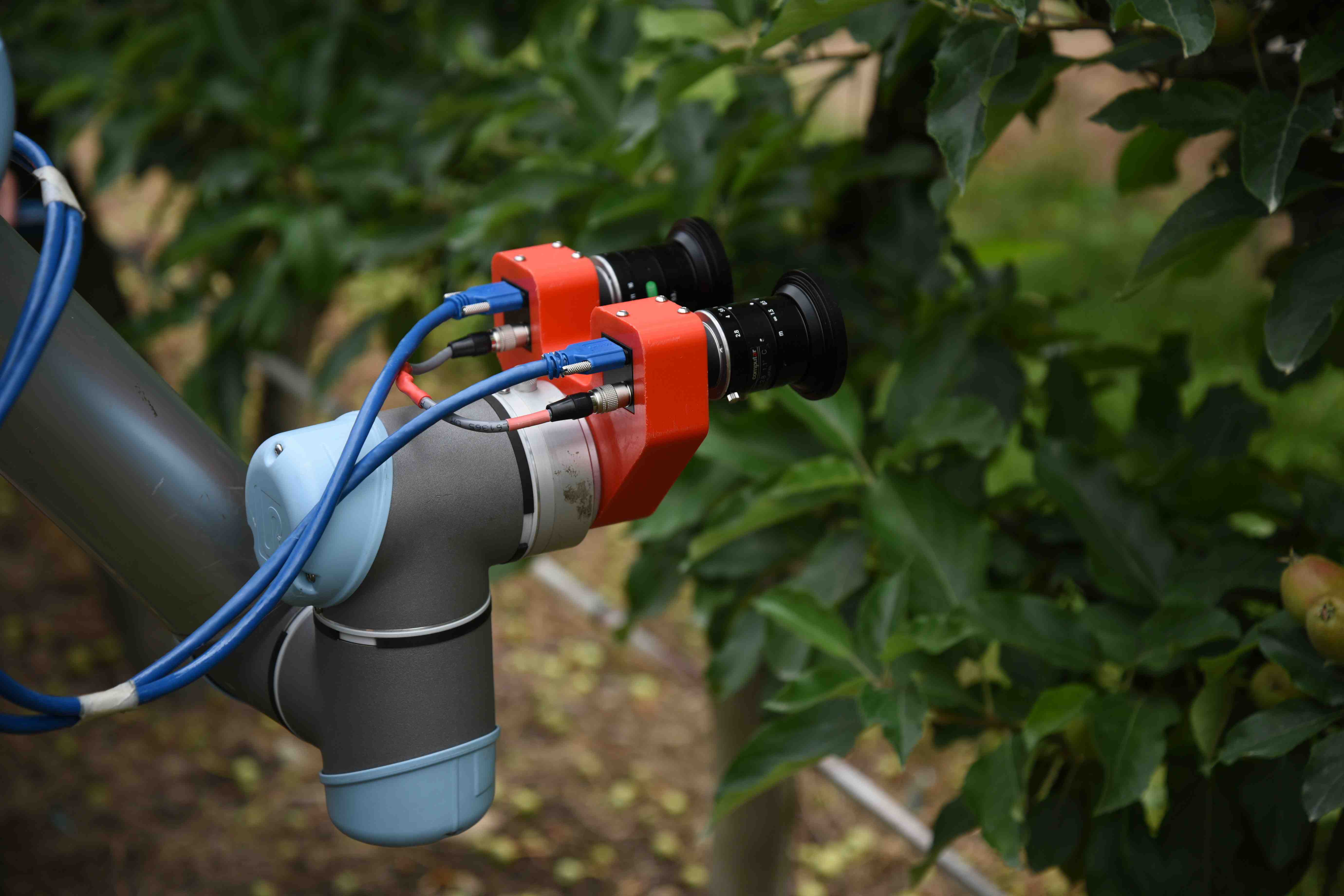}
        \caption{Stereo Cameras on the UR5 robotic arms used for scanning the apple trees (\SI{100}{\milli\meter} baseline before moving to \SI{60}{\milli\meter}).}
        \label{fig:cameras}
    \end{figure}

    The robotic platform consists of a UR5 robotic arm mounted on a simple tracked platform to move it through the orchard.
    As shown in Figure \ref{fig:cameras} stereo cameras are placed on the UR5 to scan the tree at multiple viewpoints.  
    The Blackfly BFS-U3-120S4C-CS USB 3.1 \footnote{\href{https://www.flir.com/products/blackfly-s-usb3/}{https://www.flir.com/products/blackfly-s-usb3/}} cameras were used as they have a high dynamic range suitable for dynamic outdoor lighting conditions.
    The cameras provide a high-resolution image at 4000x3000 pixels, suitable for theoretically measuring with the right calibration at millimetre accuracy.
    Computer lenses (V0828-MPY2 F2.8, 1.1" format, focal length of 8 mm) were used to provide a working distance of \SIrange{300}{600}{\milli\meter} suitable for scanning the depth of the tree and directing the thinning tool in the future.
    The cameras were spaced with a baseline of \SI{60}{\milli\meter} to minimise the stereo error for accurate depth resolution while maintaining the required field of view.
    A hardware trigger operating at \SI{20}{\hertz} was used to synchronise the cameras for accurate stereo matching.
    
\section{Vision System - Fruitlet Mapping}\label{sec:vision}
    To determine the correct thinning decisions, the vision system must produce an accurate 3D map of the fruitlets along the branch of each apple tree.
    This is a complex problem as relatively dense foliage covers the branch making the fruitlets partially or fully obscured from single viewpoints.
    Furthermore, fruitlets grow in clusters of up to five, obstructing each other.
    This section presents our novel approach to generating an accurate map of fruitlets within 2D apple tree structures.
    
    The fruitlet mapping process coordinates the UR5 arm to capture stereo-data of the tree at various pre-planned poses designed to capture a detailed view of all the fruitlets. 
    The stereo data on the platform is aligned geometrically based on hand-eye calibration, which relies on the precision of the UR5s movements and calibration between the arm and the stereo pair.
    Calibration of the stereo cameras was measured using OpenCV and a Charuco board as the calibration target.
    The hand-eye calibration between the UR5 and the stereo pair was measured using the ViSP library\footnote{\href{https://visp.inria.fr/}{https://visp.inria.fr}} using a similar approach to the work presented in \cite{antonello2017fully}.

    The specific pre-programmed scanning pattern used to collect data for this field trial was as follows.
    The arm scanned each branch in an arc at 12 points, placing the cameras at \SIrange{300}{400}{\milli\meter} from the centre of the branch, as best as could be manually aligned. 
    The arm does six arc scans spaced \SI{15}{\milli\meter} apart along the branch width. 
    A full branch scan has 72 viewpoints along \SI{900}{\milli\meter} of the branch.
    A video of this process is shown in Section \ref{sec:videos}.

\begin{figure*}
            \centering
            \includegraphics[width=\textwidth]{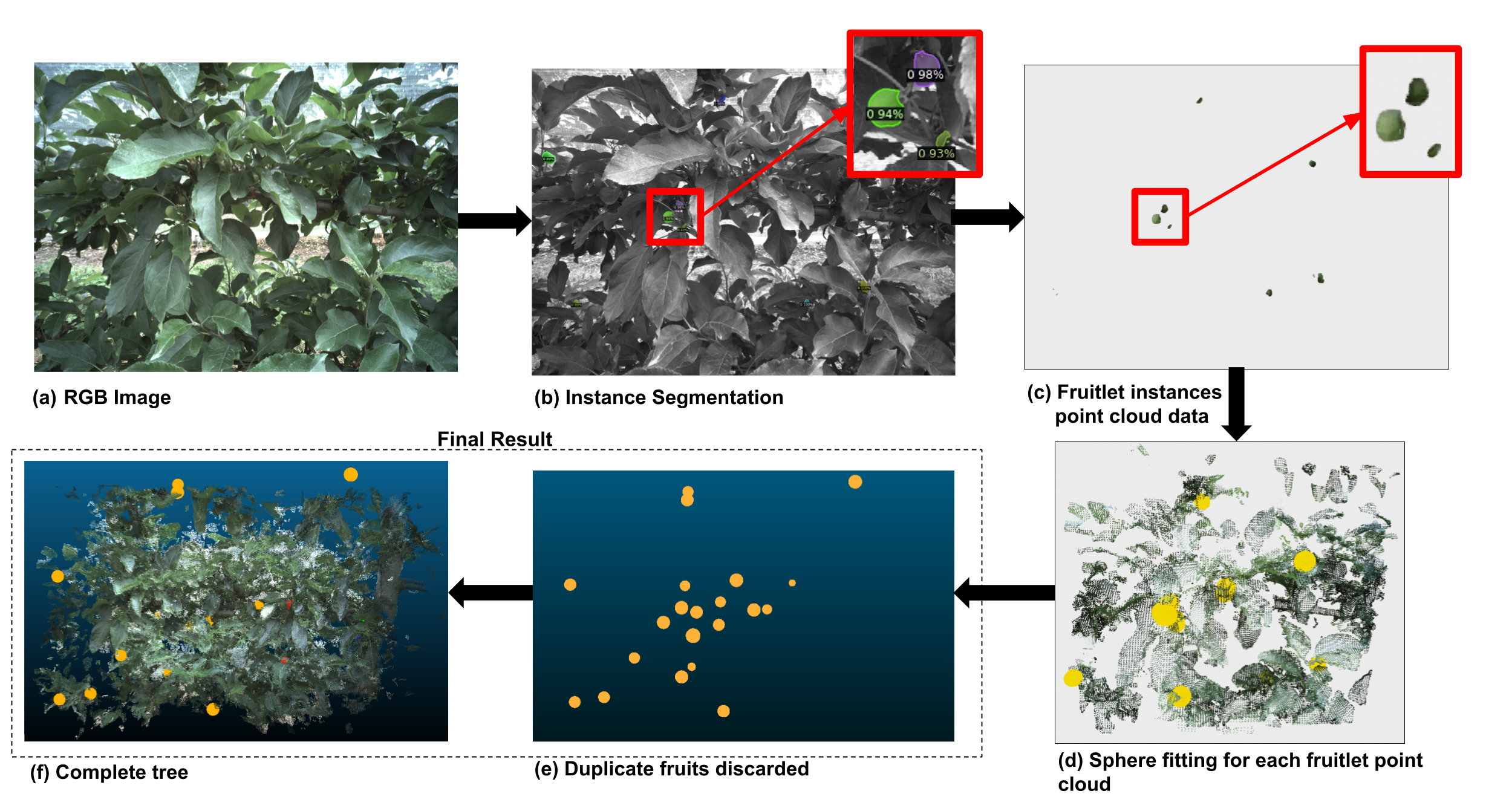}
            \caption{Visualisation of the Fruitlet Mapping Process stage by stage.}
            \label{fig:mapping}
        \end{figure*}

    The mapping process identifies the individual fruitlet at each scan point and then registers them into a 3D map as the arm scans along the branch. 
    The high-level process of processing the scans is illustrated by Figure \ref{fig:mapping}. 
    The first stage of the process extracts the fruitlet information in the immediate scan (a). 
    The instance segmentation from section \ref{sec:detection} is used to detect the frutilets within the RGB image (b). The mask is then used to extract the point cloud information for each fruitlet from the depth information generated using the stereo inference in Section \ref{sec:stereo-inference} (c). Finally, the size and shape of the fruitlets are measured and associated with the fruitlets in the overall map as in Section \ref{sec:association} (e and f).
        
    \subsection{Fruitlet Detection}\label{sec:detection}
        As has become the De facto approach for detection, our fruitlet detection leverages deep neural networks for reliable detection of the fruitlets under real-world lighting conditions.
        This step utilises Detectron2\footnote{\href{https://github.com/facebookresearch/detectron2}{https://github.com/facebookresearch/detectron2}}\cite{wu2019detectron2} as an Instance Segmentation approach to detect the fruitlets in the 2D images.

\begin{figure}[!htb]
            \centering
            \includegraphics[width=0.75\columnwidth]{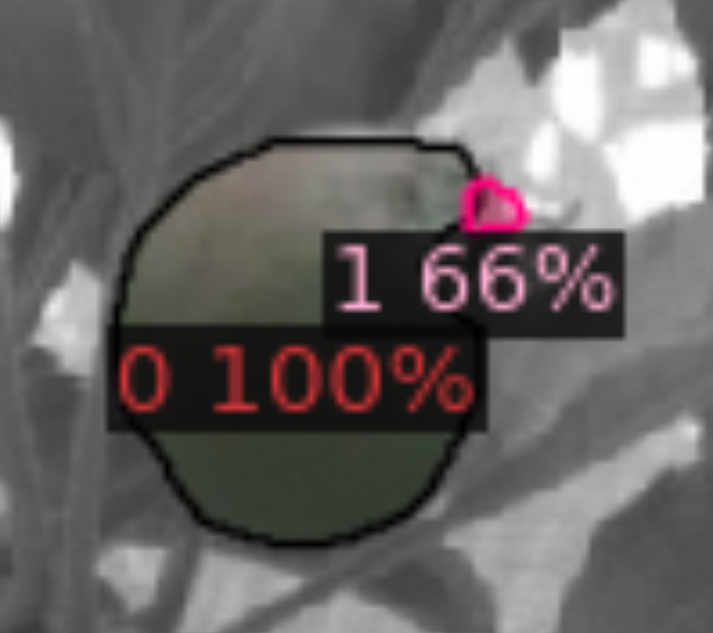}
            \caption{Example of images labelled for Instance Segmentation showing detection of the fruitlets (brown) and calyx (pink).}
            \label{fig:fruitlet-detection}
        \end{figure}

        Images were captured from several orchards throughout a variety of lighting conditions.
        70 images were manually labelled to detect \textit{fruitlets} themselves and their \textit{calyx}, and \textit{stem}.
        The calyx and stem are not used yet, future work will seek to determine the orientation of the fruitlet. 
        Figure \ref{fig:fruitlet-detection} shows an example of two of these labelled features.
        The final training performance was done using Mask R-CNN using the ResNeXT-101 backbone which gave a mean Average Precision (mAP) of 0.46.
        
    \subsection{Stereo Inference}\label{sec:stereo-inference}
        Using the tabulation of various stereo-matching benchmarks collated on the Middlebury Stereo Evaluation v3\footnote{\href{https://vision.middlebury.edu/stereo/eval3/}{https://vision.middlebury.edu/stereo/eval3/}} we explored potential approaches capable of producing low error depth maps with high-resolution images.
        Given the lack of ground truth depth data for the real-world fruitlet images we captured, we cannot train any machine learning-based networks specifically for our use case.
        
        HSMnet \cite{yang2019hierarchical} was selected for its usability on the fruitlet images based on having the lowest Mean Absolute Error in pixels Error (avgerr) for high-resolution images as of mid-2021 at $2.07$. 
        The pre-trained model was used as provided and empirically proved to produce effective depth data of the apple tree. 
        The only modification to using HSMnet was using a truncated window (around the local maxima) for doing the weighted depth regression, which we found to make it less noisy around discontinuities.
        An example of the HSMnet output on a real-world stereo image pair is shown in Figure \ref{fig:mapping}d.

    \subsection{Fruitlet Extraction}\label{sec:association}
        The primary challenge of the mapping process is correctly associating the fruitlets between scan frames to avoid duplicates.
        The challenge is to ensure that the fruit detected in one frame does not get counted again in another frame as a few bad fruitlet matches will spoil the count.
        The basic premise of the presented approach is that the centroid (centre) point of the fruitlets can be spatially aligned relative to the robot platform, presuming sufficiently accurate calibrations. 
        
        To determine the central pose of the fruitlets, we fit spheres to their associated point cloud data (Figure \ref{fig:mapping}d). Sphere fitting provides useful metrics like the centroid and radius of the fruitlet. Once the correct amount of spheres is generated for each fruitlet tree, it becomes very simple to count the spheres. However, the tricky part is ensuring that the correct spheres are generated and matched to avoid duplicates.

\begin{figure}[htb]
            \centering
            \includegraphics[width=0.8\linewidth]{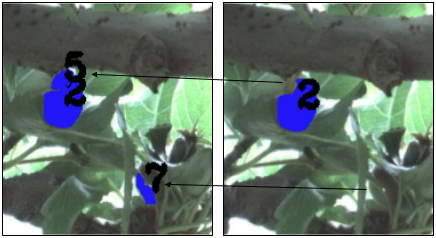}
            \caption{False Positives removed by thresholding the number of fruitlet points. }
            \label{fig:pcd_thresh}
        \end{figure}

        Incorrect counting occurs because of false positives in instance segmentation results. In many cases, leaves or small parts of the fruit would detect as a fruitlet and affect the fruitlet counting results. Since these instances were small, point cloud thresholding was done before sphere fitting to improve detection results (Figure \ref{fig:pcd_thresh}). 
        
        \subsubsection{Sphere Fitting}\label{sec:sphere-fitting}
            Different techniques such as RANSAC \cite{Fischler1981RandomSC} and Least Square Fit method \cite{pratt1987direct} can be used to fit point cloud data inside a geometric model. RANSAC fits the sphere to the point cloud by repeatedly removing the outliers. The Least Square Fit method calculates the best fit for a point cloud by minimising the sum of the offset values from the sphere curve. 

            In the case of RANSAC, a hypothetical centroid and radius are tested to find the best fit for the spherical model using Algorithm \ref{ransac}.
        
            \begin{algorithm}[h]
            \caption{Fruitlet metric extraction using RANSAC}
            \label{ransac}
            \begin{algorithmic}[1]
            	\Procedure{RANSAC}{$points$}
            	\State $i\gets 0$
            	\State $bestInliers \gets 0$
            	\State $fruitletCentroid \gets null$
            	\State $fruitletRadius \gets null$

            	\While{$i \leq iterationsMax$}
            	    \State $centroid, radius \gets Random(points)$
                	\For {$point$ in $points$}
                	    \State $dist = distance(centroid, point)$
                    	\If{$abs(dist - radius) / radius < t$}
                    	    \State $numInliers += 1$
                    	\EndIf
                	\EndFor
                	\If{$numInlers \geq bestInliers$}
                	    \State $bestInliers = numInliers$
                	    \State $fruitletCentroid = centroid$
                	    \State $fruitletRadius = radius$
                    \EndIf
                \EndWhile
            	\State Return $fruitletCentroid, fruitletRadius$
            	\EndProcedure
            \end{algorithmic}
        	\end{algorithm}

            The Least Squares Fit method utilises the equation below to calculate the centroid and radius. 
            For a point cloud with point $i$, the center \((x_{0}, y_{0}, z_{0})\) and radius \(r\) can be calculated using equation \ref{eq:sphere}:
    
            \begin{equation}
            (x_{i}-x_{0})^{2} + (y_{i}-y_{0})^{2} + (z_{i}-z_{0})^{2} = r^{2},
            \label{eq:sphere}
            \end{equation}
            for n number of points, eq \ref{eq:sphere} can be derived to:
            \begin{equation}
                \underset{\underset{f}{\rightharpoonup}}{\underbrace{\begin{bmatrix}
                     x_1^{2} + y_1^{2} + z_1^{2}\\
                    x_2^{2} + y_2^{2} + z_2^{2}\\
                     . \\
                     . \\
                     . \\
                    x_n^{2} + y_n^{2} + z_n^{2}
                    \end{bmatrix}}}=
                    \underset{A}{\underbrace{\begin{bmatrix}
                     x_1 & y_1 & z_1\\
                    x_2 & y_2 & z_2\\
                     &. \\
                     &. \\
                     &. \\
                    x_n & y_n & z_n
                    \end{bmatrix}}}*
                    \underset{\underset{c}{\rightharpoonup}}{\underbrace{\begin{bmatrix}
                     2*x_0\\
                    2*y_0\\
                    2*z_0
                    \end{bmatrix}}}.
                     \label{eq:f=Ax}
            \end{equation}
            
\begin{figure}[htb]
                \centering
                \includegraphics[width=\linewidth]{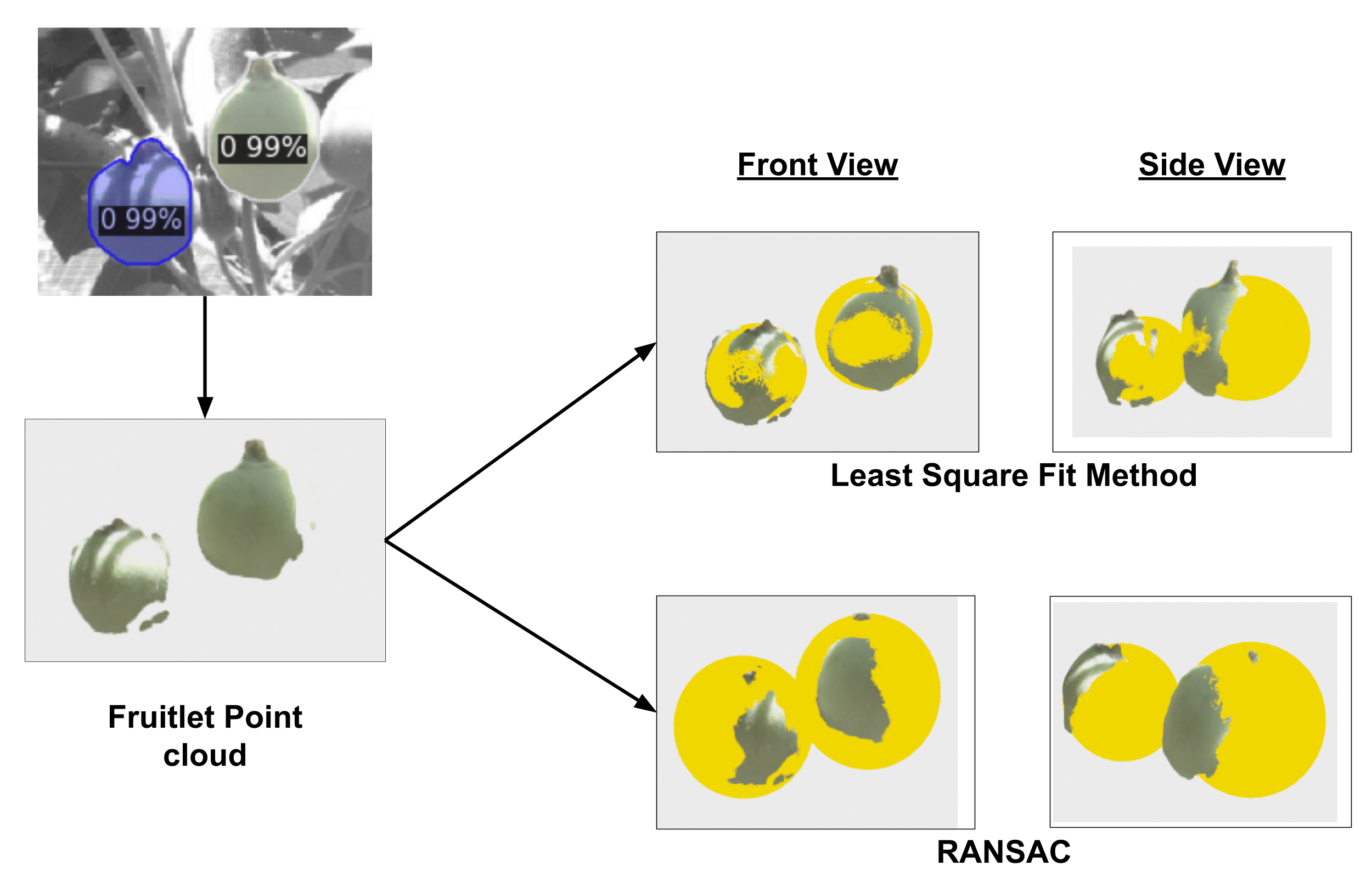}
                \caption{Example results of fitting spheres to the point cloud data using RANSAC and Least Square Fit.}
                \label{fig:fitting}
            \end{figure}

            A Euclidean 2-norm error \(\left\| f - Ac \right\|\) can be used to calculate \(c\) which gives us the centroid and radius. An example of the sphere fitting can be seen in Figure \ref{fig:fitting}. 

            \begin{figure}[htb]
                \centering
                \includegraphics[width=0.8\linewidth]{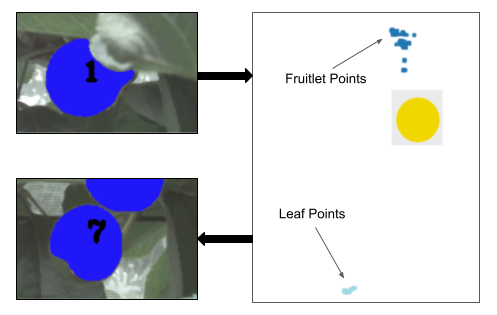}
                \caption{Centroid position is affected by occlusions which causes wrong sphere generation.}
                \label{fig:occlusion}
            \end{figure}
            
        \subsubsection{Sphere Matching}
\begin{figure}[htb]
            \centering
            \includegraphics[width=0.8\linewidth]{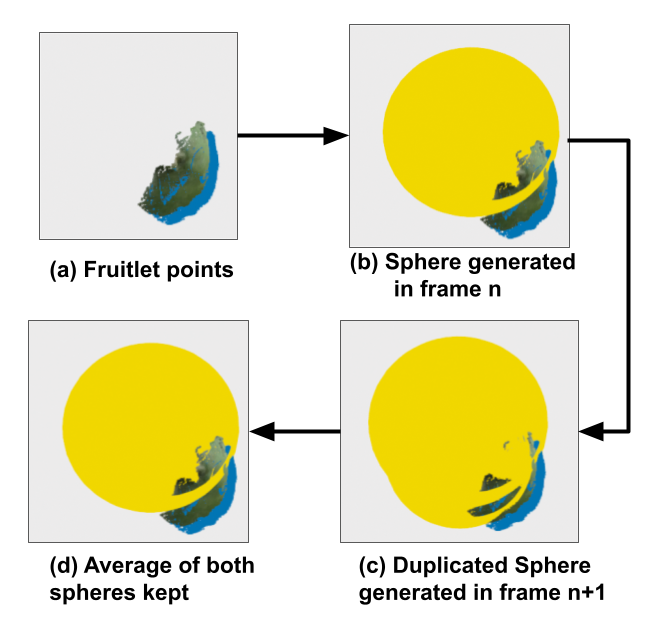}
            \caption{An example of associating fruitlets between two different viewpoints.}
            \label{fig:association}
        \end{figure}

        The spherical representation of each fruitlet needs to be further refined to avoid multi-counting. The sphere for the same fruitlet from a different angle with a similar centroid and radius can be generated again. Multi-counting would occur unless such fruitlet spherical models are thresholded properly. If the area of the newly generated sphere intersects 50\% of the area of a previously generated sphere then both would be replaced by a sphere having a centroid which is calculated by taking the average of the two centroids  (Figure. \ref{fig:association}). The same is done for the sphere radii. This is the sphere matching threshold. Increasing this threshold may affect the counting of neighbouring fruits, hence under-counting. Whereas by decreasing this threshold, the risk of over-counting increases.
        
        Instance segmentation was not always precise enough to extract the complete fruitlet boundary since there were cases of the fruitlet mask containing occlusions such as leaves or branches. An example of such a case can be seen in Figure \ref{fig:occlusion}. In the first detection, the leaf in front of the fruitlet contaminates the point cloud data and creates a depth disparity that affects the sphere's position. The leaf does not interfere in the following image, however, the new sphere generated does not intersect with the previous sphere of the same fruitlet, which creates an extra count. To reduce the effect of occlusions, the point cloud was divided using k-means clustering, and the larger part was considered the fruitlet.\\
        
\begin{figure}[htb]
                \centering
                \includegraphics[width=0.8\linewidth]{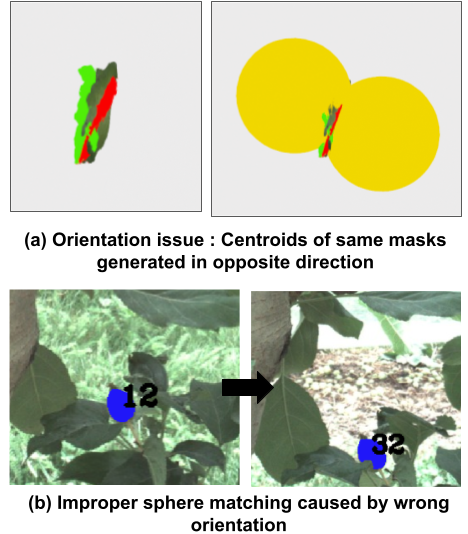}
                \caption{Multi-counting caused by centroid curvature problem.}
                \label{fig:orientation}
            \end{figure}

        Multi-counting was also caused by incorrect mask curvature. In some cases, the mask of the scans did not curve in the direction of the camera. This affected the position of the sphere created (Figure \ref{fig:orientation}a). The general idea is that the masks from the scans should always be curving away from the camera whenever sphere fitting is done, otherwise multi-counting occurs (Figure \ref{fig:orientation}b). This issue was resolved by measuring the distance of the fruitlet relative to the camera. If the fitted sphere is closer to the camera than the masked points, then the centroid direction is corrected by flipping it away from the camera.
        
        After this process, the generated spheres from figure \ref{fig:mapping}e can be added to give the total count. Each fruitlet is assigned a count number that it keeps throughout the process.

    \subsection{Video Demonstrations}\label{sec:videos}
        Videos demonstrating the scanning process and fruitlet mapping approach can be found \href{https://cares.blogs.auckland.ac.nz/research/seeing-the-fruit-for-the-leaves/}{here}\footnote{\url{https://cares.blogs.auckland.ac.nz/research/seeing-the-fruit-for-the-leaves/}}. 

\section{Experimental Setup}
    The thinning platform was evaluated in a real-world commercial apple orchard in Nelson, New Zealand, during the 2022 thinning season.
    The orchard is a mature orchard using a 2D growing system that constrains the tree to a relatively flat structure.
    The trees were not trimmed, de-leafed, or modified with the robot operating on trees as a human would be expected to.
    Data was captured over one week with the robot operating in various lighting conditions. 
    
    Data capturing was conducted by manually driving the platform in front of a branch segment, placing the scanning arm centred along the branch as shown in the videos in \ref{sec:videos}.
    The scanning process was run to capture data through the process described in Section \ref{sec:vision}.
    Once completed, the robot was driven forward to the next tree in the row, repeating as often as possible.
    Data from 34 branches were captured and evaluated, where each scan captured 28-60 RGB-D stereo image pairs.
    
\section{Results}
    The evaluation was done by manually measuring the number of fruitlets visible in the scan vs the number the system reported.
    This process does rely on a human meticulously looking through the data to determine how many fruitlets were visible to the system as it scanned.
    
    Ideally, the ground truth number of fruitlets would be measured by manually counting (or harvesting) the fruitlets along the branch. 
    However, the arm does not see the entire branch structure; it can only count the fruitlets visible within its field of view.
    Even moving the cameras around, the foliage occludes the tree's other side. 
    
    Manually counting only those within the field of view is impractical.
    Future work will utilise an arm on either side of the canopy to provide a complete view of the branch to address this. 
    Therefore, based on human observation of the results, the evaluation measures the number of fruitlets correctly identified, located, and tracked through the scan within the arm's field of view. 

    Error in the counting within the data is introduced as some fruitlets are challenging to see or track even to a human. 
    But this initial evaluation does provide a reasonable evaluation of the accuracy of the initial approach to guide future developments.
    
    For the 34 scans, the true positive, false positive, and false negative count was measured to evaluate the mapping system. The overall percentage difference was calculated to give an overall error. The percentage error will be positive in case of over-counting and negative for under-counting. The absolute percentage difference was also calculated to compare the magnitude of error found in counting using various sphere fitting techniques. 
 
    \subsection{Load Measurement}
\begin{figure}[htb]
            \centering
            \includegraphics[width=\columnwidth]{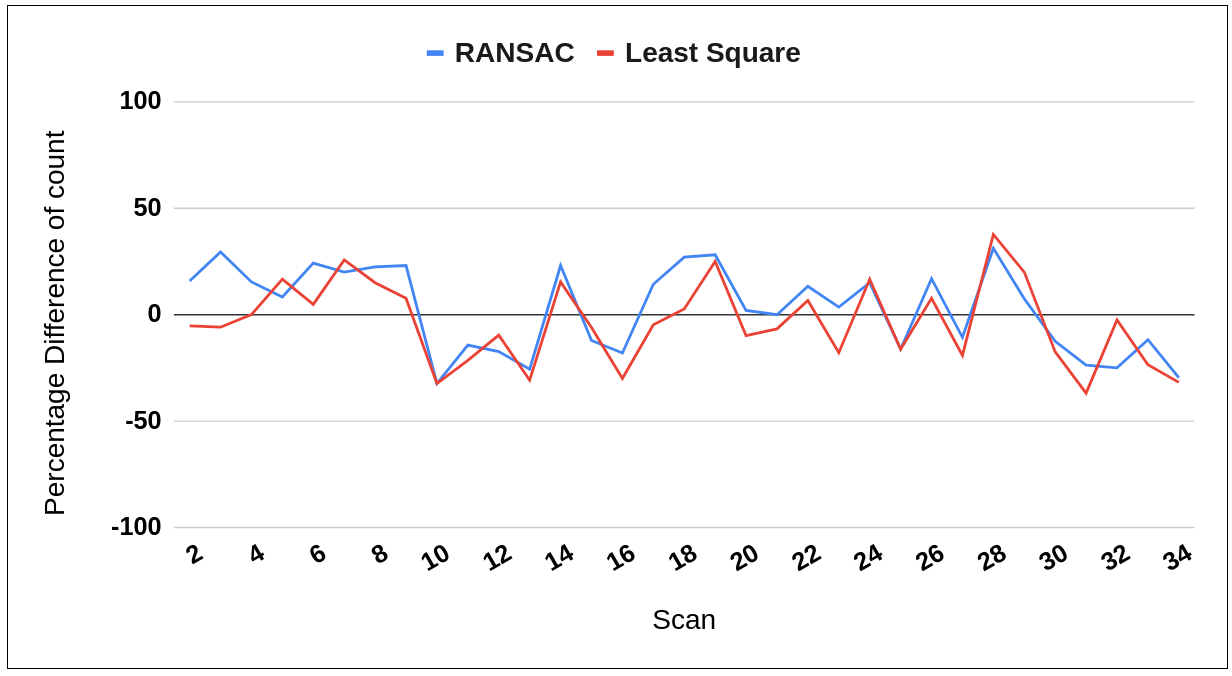}
            \caption{Graphical analysis shows that Least Square fit method curve is closer to minimum error line.}
            \label{fig:result}
        \end{figure}

        34 scans were run through the vision system described in Section \ref{sec:vision}. In the first stage, the trained Mask RCNN model with ResNeXt-101 backbone gives instances with 0.46 mAP. The resulting load predictions are compared against the human-determined ground truth in Table \ref{tab:results} and Figure \ref{fig:result}. The human-determined ground truth was counted manually by one individual and a small subset was verified by another individual.
        
        \begin{table}[]
        \centering
        \caption{Counting accuracy using different sphere fitting techniques.}
        \label{tab:results}
        \resizebox{\columnwidth}{!}{%
        \begin{tabular}{|c|cc|}
        \hline
         & \multicolumn{2}{c|}{\textit{Sphere Fitting Technique}} \\ \cline{2-3} 
        \multirow{-2}{*}{Evaluation Method} & \multicolumn{1}{c|}{RANSAC} & Least Square \\ \hline
        \cellcolor[HTML]{FFFFFF}\begin{tabular}[c]{@{}c@{}}Percentage \\ Error (\%)\end{tabular} & \multicolumn{1}{c|}{2.83} & -3.71	 \\ \hline
        \cellcolor[HTML]{FFFFFF}\begin{tabular}[c]{@{}c@{}}Absolute Percentage \\ Error (\%)\end{tabular} & \multicolumn{1}{c|}{17.48} & 15.58 \\ \hline
        \end{tabular}%
        }  	
        \end{table}
        
        From Table \ref{tab:results} it can be seen that fewer cases of over-counting occurred when the Least Square method was used for sphere fitting. RANSAC gave more cases of over-counting since the radius had a more considerable error. It can be observed that in Figure \ref{eq:sphere} the sphere is larger in size than the fruitlet surface, which affects the sphere matching accuracy and causes over-counting.
        
        The accuracy-based evaluation misses out on the over-counting and under-counting numbers. This evaluation presumes that multi-counting balances out the undetected fruitlets. A better approach would be to calculate the Precision and Recall. Table \ref{tab:eval} is generated using the sphere fitting results based on Least Square Fit method since it gave better accuracy. True Positive (TP) indicates the number of fruitlets truly counted, False Positive (FP) indicates the number of fruitlets over-counted and False Negative (FN) indicates the number of fruitlets that weren't detected through instance segmentation. For 34 data sets, a precision of 0.872 and a recall of 0.833 was achieved. In this work, the speed of the system is not focused on since the aim is to achieve counting accuracy.

\begin{table}
\centering
\caption{Counting algorithm tested on 34 individual scans}
\label{tab:eval}
\resizebox{\linewidth}{!}{%
\begin{tabular}{|c|c|c|c|c|r|r|r|} 
\hline
\begin{tabular}[c]{@{}c@{}}\textbf{Ground}\\\textbf{ }\\\textbf{ Truth}\end{tabular} & \begin{tabular}[c]{@{}c@{}}\textbf{Proposed}\\\textbf{ }\\\textbf{ Algorithm}\end{tabular} & \textbf{TP} & \textbf{FP} & \textbf{FN} & \multicolumn{1}{l|}{\textbf{Precision}} & \multicolumn{1}{l|}{\textbf{Recall}} & \multicolumn{1}{c|}{\textbf{F1 score}} \\ 
\hline
66 & 66 & 60 & 6 & 6 & 0.909 & 0.909 & 0.909 \\ 
\hline
38 & 36 & 34 & 2 & 4 & 0.944 & 0.895 & 0.919 \\ 
\hline
34 & 32 & 28 & 4 & 6 & 0.875 & 0.824 & 0.849 \\ 
\hline
39 & 39 & 35 & 4 & 4 & 0.897 & 0.897 & 0.897 \\ 
\hline
36 & 42 & 35 & 7 & 1 & 0.833 & 0.972 & 0.897 \\ 
\hline
62 & 65 & 55 & 10 & 7 & 0.846 & 0.887 & 0.866 \\ 
\hline
35 & 44 & 34 & 10 & 1 & 0.773 & 0.971 & 0.861 \\ 
\hline
40 & 46 & 39 & 7 & 1 & 0.848 & 0.975 & 0.907 \\ 
\hline
52 & 56 & 45 & 11 & 7 & 0.804 & 0.865 & 0.833 \\ 
\hline
34 & 23 & 21 & 2 & 13 & 0.913 & 0.618 & 0.737 \\ 
\hline
28 & 22 & 20 & 2 & 8 & 0.909 & 0.714 & 0.8 \\ 
\hline
52 & 47 & 41 & 6 & 11 & 0.872 & 0.788 & 0.828 \\ 
\hline
39 & 27 & 26 & 1 & 13 & 0.963 & 0.667 & 0.788 \\ 
\hline
39 & 45 & 38 & 7 & 1 & 0.844 & 0.974 & 0.904 \\ 
\hline
66 & 62 & 53 & 9 & 13 & 0.855 & 0.803 & 0.828 \\ 
\hline
50 & 35 & 33 & 2 & 17 & 0.943 & 0.66 & 0.777 \\ 
\hline
21 & 20 & 19 & 1 & 2 & 0.95 & 0.905 & 0.927 \\ 
\hline
37 & 38 & 35 & 3 & 2 & 0.921 & 0.946 & 0.933 \\ 
\hline
32 & 40 & 32 & 8 & 0 & 0.8 & 1 & 0.889 \\ 
\hline
51 & 46 & 39 & 7 & 12 & 0.848 & 0.765 & 0.804 \\ 
\hline
30 & 28 & 24 & 4 & 6 & 0.857 & 0.8 & 0.828 \\ 
\hline
45 & 48 & 41 & 7 & 4 & 0.854 & 0.911 & 0.882 \\ 
\hline
28 & 23 & 22 & 1 & 6 & 0.957 & 0.786 & 0.863 \\ 
\hline
60 & 70 & 59 & 11 & 1 & 0.843 & 0.983 & 0.908 \\ 
\hline
37 & 31 & 27 & 4 & 10 & 0.871 & 0.73 & 0.794 \\ 
\hline
65 & 70 & 64 & 6 & 1 & 0.914 & 0.985 & 0.948 \\ 
\hline
47 & 38 & 33 & 5 & 14 & 0.868 & 0.702 & 0.776 \\ 
\hline
61 & 84 & 61 & 23 & 0 & 0.726 & 1 & 0.841 \\ 
\hline
40 & 48 & 40 & 8 & 0 & 0.833 & 1 & 0.909 \\ 
\hline
40 & 33 & 24 & 9 & 16 & 0.727 & 0.6 & 0.657 \\ 
\hline
38 & 24 & 20 & 4 & 18 & 0.833 & 0.526 & 0.645 \\ 
\hline
40 & 39 & 35 & 4 & 5 & 0.897 & 0.875 & 0.886 \\ 
\hline
34 & 26 & 25 & 1 & 9 & 0.962 & 0.735 & 0.833 \\ 
\hline
44 & 30 & 29 & 1 & 15 & 0.967 & 0.659 & 0.784 \\ 
\hline
\multicolumn{5}{|r|}{\textbf{Average Values:}} & \multicolumn{1}{c|}{\textbf{0.872}} & \multicolumn{1}{c|}{\textbf{0.833}} & \multicolumn{1}{c|}{\textbf{0.844}} \\
\hline
\end{tabular}
}
\end{table}
        \section{Discussion}
            A few bad apples can spoil the count. Bad detection causes under-counting whereas bad sphere matching causes over-counting. 3D information and a multi-view approach helped in overcoming the occlusion problem to an extent. The fruitlets that are hidden from one view are detected in another view. With each frame, the 2D and 3D information needs to be associated and matched properly. Sphere fitting was found to be an effective approach for the fruitlet association when good depth information was obtained. Sphere size-based thresholding and the various other fruitlet matching methods reduced the over-counting problem but did cause under-counting.

            Qualitatively, we saw a number of examples of poor depth data due to minor obstructions from the canopy blocking part of the fruit (leaves, canes etc). In some cases this was the only observation of that specific fruitlet, meaning if it was filtered then it would not be counted. Increasing the density of the scans may overcome this issue without relaxing the filtering. By observing the fruitlet from more angles we increase the chance of obtaining clear depth information about the fruit which leads to effective associations. This could enable the system to filter out the poorer observations further and improve the sphere matching and fitting. 
            
            The significant factors that affect fruitlet counting are the quality of the instance detection model and the depth data. A less accurate instance segmentation will create undercounting, whereas a less precise instance segmentation will cause overcounting. If the quality of the depth data is poor then the association between 2D detection and 3D data can be inaccurate. In such cases, factors such as the sphere fitting techniques and the sphere matching threshold become important. Other factors include the clustering distance, the minimum number of fruitlet points allowed, and the instance segmentation score threshold. Tuning these parameters can affect the results. A balance needs to be created between these parameters to minimise both over-counting and under-counting. This also deduced that the fruitlet radius size matters a lot in load measurement and a precise size calculation algorithm is necessary for precise fruitlet counting. 

\section{Conclusions and Future Work}
    In this proposal, we have presented the initial design of the vision system for an automated apple fruitlet thinning system. Unlike previous yield measuring systems that estimate the overall load, we have specifically attempted to map the fruitlets along the individual branches of the 2D tree structures. We evaluated the vision system on 34 scans of apple trees in a commercial orchard under real-world conditions. The results demonstrate that the platform is capable of measuring the load of an apple tree with an accuracy of 84.42\% and a precision of 87.2\%.

    Future work is still required to reduce the under-counting through improvements to the detection system, and further developments to the sphere fitting approach. An evaluation of the size estimates against a ground truth measurement is also still required. To get effective fruitlet thinning results, this fruitlet counting algorithm combined with size estimation and quality assessment fruitlets can be useful.

\section*{Acknowledgements}
    This research was supported by the New Zealand Ministry for Business, Innovation and Employment (MBIE) on contract UOAX1810.

\bibliography{ref}
\bibliographystyle{named}
\end{document}